# Using Alternation Direction Method of Multipliers to Enhance robots Calibration Accuracy based on Multi-Planal Constraints

Tinghui Chen, Shuai Li, *Senior Member, IEEE*

*Abstract*—With the widespread application of industrial robots, the problem of absolute positioning accuracy becomes increasingly prominent. To ensure the working state of the robots, researchers commonly adopt calibration techniques to improve its accuracy. However, an industrial robot's working space is mostly restricted in real working environments, making the collected samples fail in covering the actual working space to result in the overall migration data. To address this vital issue, this work proposes a novel industrial robot calibrator that integrates a measurement configurations selection (MCS) method and an alternation-direction-method-of-multipliers with multiple planes constraints (AMPC) algorithm into its working process, whose ideas are three-fold: a) selecting a group of optimal measurement configurations based on the observability index to suppress the measurement noises, b) developing an AMPC algorithm that evidently enhances the calibration accuracy and suppresses the long-tail convergence, and c) proposing an industrial robot calibration algorithm that incorporates MCS and AMPC to optimize an industrial robot's kinematic parameters efficiently. For validating its performance, a public-available dataset (HRS-P) is established on an HRS-JR680 industrial robot. Extensive experimental results demonstrate that the proposed calibrator outperforms several state-of-the-art models in calibration accuracy.

## I. Introduction

TO Date, an industrial robot is a complex intelligent agent integrating mechanical, electronic, computer, control, sensor and other multidisciplinary technologies, which is an indicator to measure the industrial development level of a country [1]-[4]. Generally, an industrial robot enjoys towering repetitive positioning accuracy, while the absolute positioning accuracy is frequently low, which greatly restricts the application of robots in high-precision fields [5][9].

The most significant technique to enhance the absolute positioning accuracy of an industrial robot is kinematic parameters calibration [10]-[14]. This technique involves sufficient samples to cover the entire workspace of the industrial robot. However, in the actual working environment, many robots' workspaces are partially occluded causing the lack of some samples, which affects data integrity and the final calibration results. To address this tough issue, scholars develop various constraint methods for samples collection. Lembono *et al*. [15] employ a laser and a tooltip to achieve the planar constraint and the distance constraint, which can achieve high-accuracy robot calibration. Wang *et al*. [16] propose a kinematic calibration method based on the point constraint and the distance constraint for enhancing the absolute position accuracy of a 6R robot.

However, the above algorithms all adopt conventional algorithms for the kinematic parameters calibration like the LM algorithm, the Gauss-Newton algorithm and the LS (least-square) algorithm [16]-[22]. These algorithms frequently encounter long-tail convergence, which decreases computational efficiency [23]-[37]. Hence, motivated by the above analysis, we propose a novel alternation direction method of multipliers (ADMM)-based multi-planal constraints (MPC) model (AMPC) with the following two-fold ideas:

a) establishing an efficient industrial robot calibration scheme based on the ADMM principle, to achieve a fast convergence and a high calibration accuracy.
b) Incorporating multi-planal constraints into the ADMM algorithm for addressing the mentioned issue that the robot's workspaces partially are occluded resulting in the lack of some samples.

The observability indexes are employed as the evaluation metrics to suppress the measurement noise. In general, the larger observability index indicates less measurement noise and stable identification performance [38]. This paper adopts DE (differential evolution) algorithm to optima observability indexes for the measurement configurations selection (MCS).

Therefore, the main contributions of this work include:

a) An industrial robot calibration with MPC dataset HSR-P. It concludes the data of six joint angles obtained from the teach pendant of an HSR-JR680 industrial robot, the length of the drawstrings and a dial indicator indication. It is exposed on the website (https://github.com/chenth08/HSR-P) to bring data assistance for scholars;
b) An AMPC algorithm. It identifies the kinematic parameters with high effectiveness. It incorporates multi-planal constraints into the ADMM algorithm to ensure its high calibration accuracy and fast convergence; and
c) MCS-AMPC algorithm. It can further enhance the calibration accuracy of the industrial robot.

Section II expounds on the preliminaries. Section III gives the method of kinematic parameters identification. For Section IV, we report the experimental results. Finally, Section V illustrates the conclusion.

## II. Preliminaries

For a D-H model [39]-[39], which contains link length $a_i$, joint angle $\theta_i$, link offset $d_i$, link twist angle $\alpha_i$. Table I lists the initial D-H parameters of an HSR-JR680 robot, whose transformation matrix between the adjacent links is given as:

TABLE I. ABB IRB120 Industrial robot D-H parameters.

| Joint $i$ | $\alpha_i/°$ | $a_i/mm$ | $d_i/mm$ | $\theta_i/°$ |
|---|---|---|---|---|
| 1 | -90 | 250 | 653.5 | 0 |
| 2 | 0 | 900 | 0 | -90 |
| 3 | 90 | -205 | 0 | 180 |
| 4 | -90 | 0 | 1030.2 | 0 |
| 5 | 90 | 0 | 0 | 90 |
| 6 | 0 | 0 | 200.6 | 0 |

$$\Lambda_i = Rot_Z(\theta_i) \times Trans_Z(d_i) \times Trans_X(a_i) \times Rot_X(\alpha_i) = \begin{bmatrix} \cos\theta_i & -\sin\theta_i c\alpha_i & \sin\theta_i s\alpha_i & a_i\cos\theta_i \\ \sin\theta_i & \cos\theta_i c\alpha_i & -\cos\theta_i s\alpha_i & a_i\sin\theta_i \\ 0 & \sin\alpha_i & \cos\alpha_i & d_i \\ 0 & 0 & 0 & 1 \end{bmatrix}, \quad (1)$$

where $\Lambda_i$ refers to the transformation matrix [40]-[45]. Then, based on (1), the kinematic model of a robot is given as:

$$\Lambda = \Lambda_1\Lambda_2\Lambda_3\Lambda_4\Lambda_5\Lambda_6, \quad (2)$$

where $H_N$ is the nominal matrix, and $\hat{P}_N$ refers to the robot's nominal position vector. Owning to the errors of kinematic parameters, the correspondingly actual pose error matrix of an industrial robot end-effector can be expressed as:

$$\Delta\Lambda = \Lambda_r - \Lambda = \begin{bmatrix} \Delta H & \Delta\hat{P} \\ 0 & 1 \end{bmatrix}, \quad (3)$$

where, $\Lambda_r$ is the actual pose matrix of an industrial robot end-effector, and $\Lambda$ refers to the nominal pose matrix of an industrial robot end-effector.

Moreover, the $\Lambda_i$'s holomorphic differential form is given as:

$$d\Lambda_i = \frac{\partial \Lambda_i}{\partial \alpha_i}\Delta\alpha_i + \frac{\partial \Lambda_i}{\partial a_i}\Delta a_i + \frac{\partial \Lambda_i}{\partial d_i}\Delta d_i + \frac{\partial \Lambda_i}{\partial \theta_i}\Delta\theta_i = \Lambda_i \delta\Lambda_i. \quad (4)$$

Then, the pose error model can be established as:

$$\delta E = \begin{bmatrix} J_1 & J_2 & J_3 & J_4 \end{bmatrix}\begin{bmatrix} \Delta\alpha & \Delta a & \Delta d & \Delta\theta \end{bmatrix}^T = J_r \delta\chi, \quad (5)$$

where $J_r$ is the Jacobian matrix of the error model, $\Delta\chi$ is the kinematic parameters error vector containing $\Delta\theta$, $\Delta a$, $\Delta d_i$, $\Delta\alpha$.

III. KINEMATIC PARAMETERS IDENTIFICATION

*A. Measurement Configurations Selection (MCS)*

The process for measurement configurations selection (MCS) is to investigate a group of poses in an industrial robot workspace for minimizing the impact of noise on the estimation of robot kinematics parameters [38]. With $N_0$ different measuring configurations, (5) can be expanded as:

$$\Delta E = \hat{J}\Delta\chi, \quad (6)$$

where $\hat{J}=[J_{r1}, J_{r2},…, J_{rN0}]$ represents error extended Jacobian matrix at $N_0$ initial configurations, Then, based on SVD (Singular value decomposition), (8) can be depicted as:

$$\Delta E = G\Sigma Q^T \Delta\chi, \quad (7)$$

where $G$ and $Q$ represent orthogonal matrices, $\Sigma$ refers to the diagonal matrix containing $m$ singular values ($\sigma_1, \sigma_2, \cdots, \sigma_m$), which are conformed to $\sigma_1 \geq \sigma_2 \geq \cdots \geq \sigma_m \geq 0$.

To address the issue of the MCS, Wang *et. al.* propose a new observability index for the industrial robot kinematic calibration, which is given as [38]:

$$O_6 = \left[\left(\frac{1}{N_0}\sum_{i=1}^{N_0}\left(\frac{1}{\sigma_i^2}\right)^{\frac{V}{2}}\right)\right]^{-\frac{1}{V}}, \quad (8)$$

where $z$ refers to the number of kinematic parameters. $V$ is a parameter, which is set as 2 according to [38]. Note that the larger observability indexes indicate less measurement noise and stable identification performance.

*B. AMPC Algorithm*

Since the nonlinear characteristics of the D-H model (1), the regularization terms are added to avoid the issues of over-fitting. Moreover, according to the principle of ADMM [46], to satisfy the multi-planal constraints, we build the following augmented Lagrange function:

$$\begin{aligned}
\arg\min f = &\frac{1}{2nN}\sum_{j=1}^{n}\sum_{i=1}^{N}\left\|L_{i,j}-\hat{L}_j-\frac{\partial \hat{L}_{i,j}}{\partial u_t}(u_{t+1}-u_t)-\frac{\partial \hat{L}_{i,j}}{\partial P_{0,t}}(P_{t+1}-P_t)\right\|_2^2 \\
&+\frac{1}{nN}\sum_{j=1}^{n}\sum_{i=1}^{N}\left\langle \Phi_{i,j}(u_t,W_{j,t},\gamma_{j,t})+\frac{\partial \Phi_{i,j}}{\partial u_t}(u_{t+1}-u_t)+\frac{\partial \Phi_{i,j}}{\partial W_{j,t}}(W_{j,t+1}-W_{j,t})+\frac{\partial \Phi_{i,j}}{\partial \gamma_{j,t}}(\gamma_{j,t+1}-\gamma_{j,t}),\Gamma_j\right\rangle \\
&+\frac{1}{2nN}\sum_{j=1}^{n}\sum_{i=1}^{N}\rho_j\left\|\Phi_{i,j}(u_t,W_{j,t},\gamma_{j,t})+\frac{\partial \Phi_{i,j}}{\partial u_t}(u_{t+1}-u_t)+\frac{\partial \Phi_{i,j}}{\partial W_{j,t}}(W_{j,t+1}-W_{j,t})+\frac{\partial \Phi_{i,j}}{\partial \gamma_{j,t}}(\gamma_{j,t+1}-\gamma_{j,t})\right\|_2^2 \\
&+\frac{1}{2n}\sum_{j=1}^{n}\lambda_j\left[(u_{t+1}-u_t)^2+(P_{t+1}-P)^2+\sum_{j=1}^{n}(W_{j,t+1}-W_{j,t})^2+(\gamma_{j,t+1}-\gamma_{j,t})^2\right],
\end{aligned} \quad (9)$$

where $I$ is a unit matrix, $\Phi_j$ is the constraint equation of the plane $j$, $\gamma_j$ is the normal vector of the plane $j$, $W_j$ is a point of plane $j$, $n$ represents the number of planes, and this work set $n$ as 3, $\lambda_j$ is a regularized coefficient for a substantial term of the plane $j$. $\Gamma_{j,t}$ refers to the Lagrange multiplier of the plane $j$, $\rho_j$ is the constraint coefficient of the plane $j$, $\langle \cdot \rangle$ refers to the inner product between two involved matrices.

With (9), we update the $W_{j,t}$, $\gamma_{j,t}$, $u_{j,t}$ and $\Gamma_{j,t}$ as:

$$W_{j,t+1}=W_{j,t}-\left[\frac{1}{N}\sum_{i=1}^{N}\left(\rho_j\frac{\partial \Phi_{i,j,t}}{\partial W_{j,t}}\left(\frac{\partial \Phi_{i,j,t}}{\partial W_{j,t}}\right)^T\right)+\lambda_j I\right]^{-1}\times\frac{1}{N}\sum_{i=1}^{N}\left(\rho_j\Phi_{i,j}\frac{\partial \Phi_{i,j,t}}{\partial W_{j,t}}+\Gamma_j\frac{\partial \Phi_{i,j,t}}{\partial W_{j,t}}\right), \quad (10)$$

$$\gamma_{j,t+1}=\gamma_{j,t}-\left[\frac{1}{N}\sum_{i=1}^{N}\left(\rho_j\frac{\partial \Phi_{i,j,t}}{\partial \gamma_{j,t}}\left(\frac{\partial \Phi_{i,j,t}}{\partial \gamma_{j,t}}\right)^T\right)+\lambda_j I\right]^{-1}\times\frac{1}{n}\sum_{i=1}^{N}\left(\rho_j\Phi_{i,j}\frac{\partial \Phi_{i,j,t}}{\partial \gamma_{j,t}}+\Gamma_j\frac{\partial \Phi_{i,j,t}}{\partial \gamma_{j,t}}\right), \quad (11)$$

$$u_{t+1}=u_t+\frac{1}{nN}\sum_{j=1}^{n}\left[\sum_{i=1}^{N}\left(\frac{\partial \hat{L}_j}{\partial u_t}\left(\frac{\partial \hat{L}_j}{\partial u_t}\right)^T+\rho_j\frac{\partial \Phi_{i,j,t}}{\partial u_t}\left(\frac{\partial \Phi_{i,j,t}}{\partial u_t}\right)^T\right)+\lambda_j I\right]^{-1}\times\frac{1}{nN}\sum_{j=1}^{n}\sum_{i=1}^{N}\left[\frac{\partial \hat{L}_i}{\partial u_i}(L_{i,j}-\hat{L}_j)+\rho_j\Phi_{i,j,t}\frac{\partial \Phi_{i,j,t}}{\partial u_t}+\Gamma_j\frac{\partial \Phi_{i,j,t}}{\partial u_t}\right], \quad (12)$$

$$\Gamma_{j,t+1}=\Gamma_{j,t}+\frac{\eta_j\rho_j}{N}\sum_{i=1}^{N}\Phi_{i,j,t}, \quad (13)$$

where $\eta_j$ is the learning rate of the Lagrange multiplier.

IV. EXPERIMENTS AND RESULTS

A. General Settings

**Evaluation Metrics.** This paper employs the following evaluation metrics, e.g., maximum error (Max), average error (Mean) and mean square error (RMSE) [47]-[52]:

$$MAX = max\left(\sqrt{(L_i-L_i')^2}\right), i=1,2,...,N$$

$$STD = \frac{1}{N}\sum_{i=1}^{N}\left(\sqrt{(L_i-L_i')^2}\right),$$

$$RMSE = \sum_{i=1}^{N}\left(\sqrt{\frac{1}{N}(L_i-L_i')^2}\right).$$

**Datasets.** For verifying the performance of the proposed method, we adopt 3 datasets collected from an HSR-JR680 industrial robot and three planes (HSR-P1, HRS-P2 and HRS-P3). Each dataset contains 800 samples collected from a plane parallel to the ground, a plane at an angle of 45 degrees and a plane that to perpendicular to the ground. Note that each sample in D1-3 contains six joint angles, a corresponding cable length and a reading of the dial indicator. We combine the HSR-P1, HRS-P2 and HRS-P3 into HRS-P.

To enhance the computational efficiency, we randomly draw 200 samples from each dataset to build three smaller datasets (D'1-3). Note that, to verify the impact of multi-planal constraints in our experiments, we redefine three datasets H1-3, where H1, H2 and H3 contain D'1, D'1-2 and D'1-3, respectively. Then, we divide each selected dataset into the training set and test set by 2:8. Additionally, for reducing the biases coursing by data splitting, we conduct each experiment 10 times, which are illustrated with the standard deviations.

**Experimental System.** Fig. 1 depicts the experimental system for the industrial robot accuracy calibration contains a 6-DOF HSR-JR680 industrial robot, a drawstring displacement sensor, a drawstring displacement indicator, a dial indicator, an RS485 communication module and a PC.

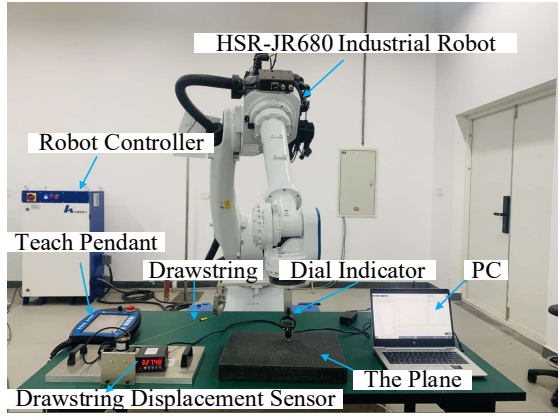

Fig. 1. The experimental system.

*Experimental Process.* We collect the samples on three different planes of various measurement positions. And, the lengths of the drawstring and dial indicator readings are recorded by the data acquisition software in real time. Subsequently, we select some samples as measurement configurations. Based on them, the AMPC algorithm is employed to identify kinematic parameter errors. Lastly, we compensate for the errors to achieve increased calibration accuracy of the industrial robot.

### B. Comparisons

*Compared Methods.* In this section, we compare the calibration performance of eight state-of-the-art algorithms against the proposed MCS-AMPC algorithm for its experimental performance verification. All involved algorithms are summarized in Table II.

TABLE II. Compared algorithms.

| Model | Description |
|---|---|
| M1 | The particle swarm optimization (PSO)-based MPC algorithm. The PSO is inspired by the movement of birds [27]. |
| M2 | The extended Kalman filter (EKF)-based MPC algorithm. The EKF is generally adopted to dispose of the issue with non-Gauss noise [5], [10]. |
| M3 | The Levenberg-Marquardt (LM)-based MPC algorithm, which is introduced in [1], [37]. |
| M4 | The simulate anneal (SA)-based MPC algorithm in [38] |
| M5 | The Unscented Kalman filter (UKF)-based MPC algorithm is introduced in [39]. |
| M6 | The least squares (LS)-based MPC algorithm in [33]. |
| M7 | The particle filter (PF)-based MPC algorithm. The PF can reduce noises as mentioned in [5]. |
| M8 | The AMPC algorithm is introduced in Section III(B). |
| M9 | The proposed MCS-AMPC algorithm. |

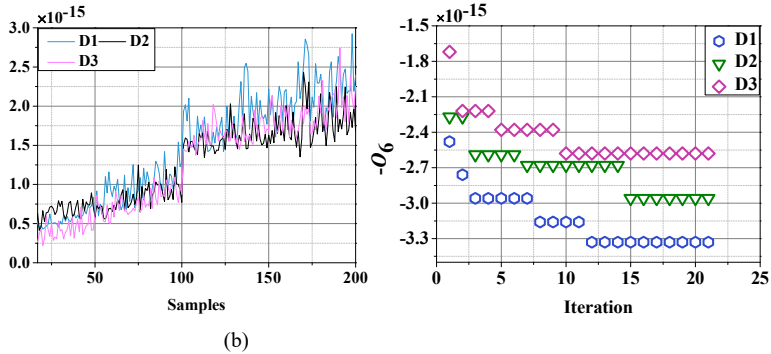

(a)          (b)

Fig. 2. Measurement configurations selection. (a) The number of measurement configurations affecting $O_6$. (b) The optimal $O_6$ for searching the best measurement configurations.

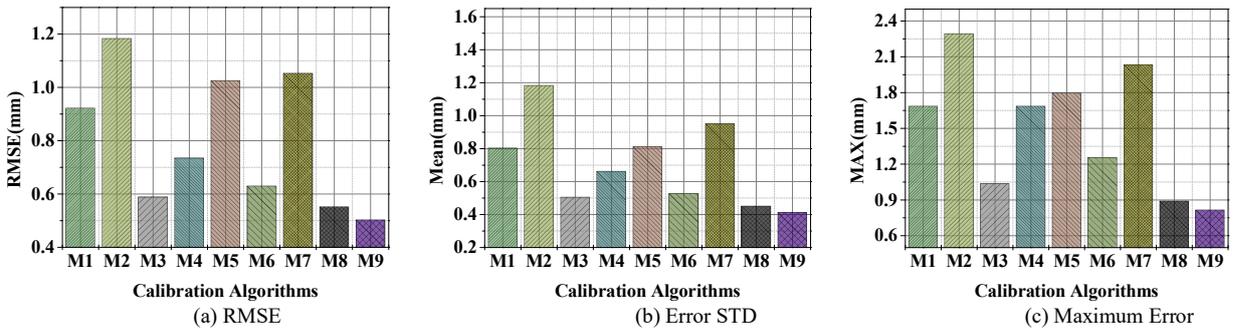

(a) RMSE      (b) Error STD      (c) Maximum Error

Fig. 3. Calibration accuracy of M1-9 on H3.

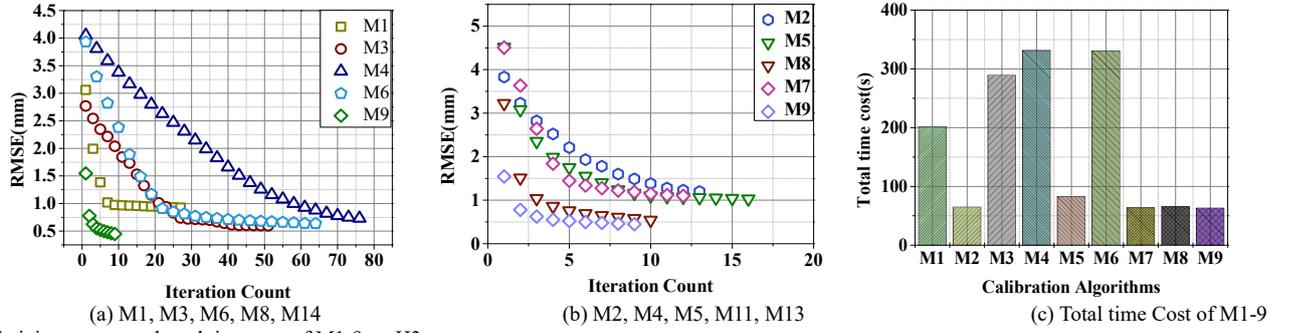

Fig. 4. Training curves and total time cost of M1-9 on H3.

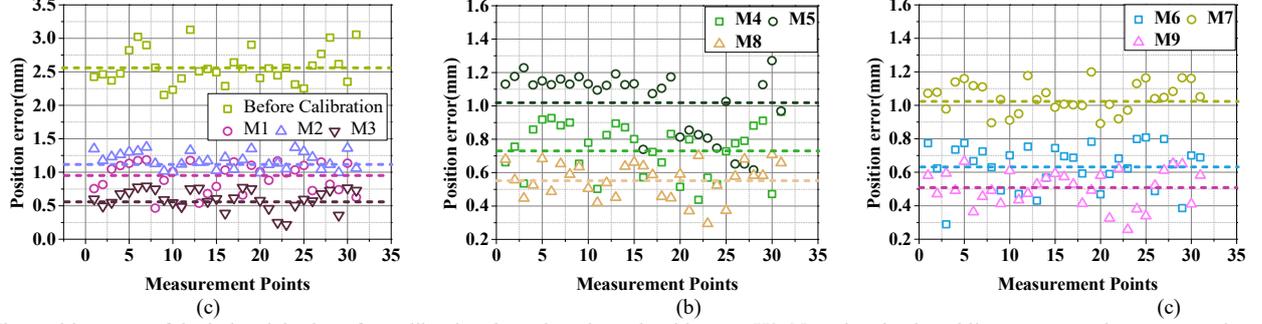

Fig. 5. The position error of the industrial robot after calibrating through various algorithms on H3. Note that the dotted lines represent the average values. It illustrates that M9 has the best calibration accuracy.

TABLE III. CALIBRATION ACCURACY OF THE COMPARED ALGORITHMS ON H1-3.

| Algorithm | H1 | | | H2 | | | H3 | | |
|---|---|---|---|---|---|---|---|---|---|
| | RMSE/mm | STD/mm | MAX/mm | RMSE/mm | STD/mm | MAX/mm | RMSE/mm | STD/mm | MAX/mm |
| Before | 2.56 | 2.45 | 4.51 | 2.56 | 2.45 | 4.51 | 2.56 | 2.45 | 4.51 |
| M1 | $0.982_{\pm6.0E-2}$ | $0.876_{\pm2.2E-2}$ | $1.756_{\pm4.9E-2}$ | $0.931_{\pm2.1E-2}$ | $0.822_{\pm1.8E-2}$ | $1.702_{\pm4.7E-2}$ | $0.922_{\pm3.1E-2}$ | $0.803_{\pm3.7E-2}$ | $1.686_{\pm5.1E-2}$ |
| M2 | $1.253_{\pm3.3E-2}$ | $1.165_{\pm3.8E-2}$ | $2.426_{\pm6.5E-2}$ | $1.196_{\pm1.1E-2}$ | $1.107_{\pm9.5E-3}$ | $2.375_{\pm3.0E-2}$ | $1.183_{\pm5.1E-2}$ | $1.066_{\pm4.0E-2}$ | $2.293_{\pm6.6E-2}$ |
| M3 | $0.655_{\pm0.0E-0}$ | $0.575_{\pm0.0E-0}$ | $1.126_{\pm0.0E-0}$ | $0.602_{\pm7.4E-2}$ | $0.520_{\pm7.4E-2}$ | $1.940_{\pm5.1E-2}$ | $0.589_{\pm7.3E-2}$ | $0.506_{\pm6.1E-2}$ | $1.037_{\pm8.2E-2}$ |
| M4 | $0.815_{\pm6.7E-3}$ | $0.723_{\pm6.0E-3}$ | $1.751_{\pm1.8E-2}$ | $0.753_{\pm3.3E-2}$ | $0.683_{\pm4.6E-2}$ | $1.700_{\pm8.7E-2}$ | $0.735_{\pm5.5E-2}$ | $0.662_{\pm3.2E-2}$ | $1.686_{\pm65E-2}$ |
| M5 | $1.080_{\pm5.6E-2}$ | $0.876_{\pm5.0E-2}$ | $1.860_{\pm5.3E-2}$ | $1.032_{\pm3.5E-2}$ | $0.820_{\pm6.5E-2}$ | $1.811_{\pm9.5E-3}$ | $1.025_{\pm3.7E-2}$ | $0.812_{\pm2.9E-2}$ | $1.796_{\pm7.3E-2}$ |
| M6 | $0.699_{\pm0.0E-0}$ | $0.598_{\pm0.0E-0}$ | $1.322_{\pm0.0E-0}$ | $0.645_{\pm6.7E-3}$ | $0.543_{\pm6.0E-3}$ | $1.279_{\pm1.8E-2}$ | $0.630_{\pm1.1E-2}$ | $0.526_{\pm8.5E-3}$ | $1.256_{\pm1.2E-2}$ |
| M7 | $1.112_{\pm6.5E-2}$ | $1.016_{\pm5.6E-2}$ | $2.102_{\pm7.6E-2}$ | $1.062_{\pm6.0E-2}$ | $0.966_{\pm2.2E-2}$ | $2.054_{\pm4.9E-2}$ | $1.053_{\pm5.3E-2}$ | $0.951_{\pm2.5E-2}$ | $2.033_{\pm7.3E-2}$ |
| M8 | $0.597_{\pm2.3E-2}$ | $0.492_{\pm2.1E-2}$ | $0.967_{\pm2.6E-2}$ | $0.568_{\pm2.6E-2}$ | $0.469_{\pm1.6E-2}$ | $0.913_{\pm3.10E-2}$ | $0.552_{\pm6.1E-2}$ | $0.452_{\pm3.1E-2}$ | $0.891_{\pm6.1E-2}$ |
| M9 | $0.550_{\pm2.3E-2}$ | $0.452_{\pm2.1E-2}$ | $0.886_{\pm2.6E-2}$ | $0.529_{\pm0.0E-0}$ | $0.425_{\pm0.0E-0}$ | $0.832_{\pm0.0E-0}$ | $0.503_{\pm0.0E-0}$ | $0.412_{\pm0.0E-0}$ | $0.816_{\pm0.0E-0}$ |

TABLE IV. THE TIME COSTS OF ALGORITHMS M1-9 ON H1-3.

| Dataset | Item | M1 | M2 | M3 | M4 | M5 | M6 | M7 | M8 | M9 |
|---|---|---|---|---|---|---|---|---|---|---|
| H1 | Iteration | $26.2_{\pm1.52}$ | $12.5_{\pm1.23}$ | $50.0_{\pm0.00}$ | $77.3_{\pm3.05}$ | $16.1_{\pm1.10}$ | $65.0_{\pm1.00}$ | $13.0_{\pm0.08}$ | $11.78_{\pm0.00}$ | $2_{\pm0.00}$ |
| | Time/s | $75.8_{\pm0.72}$ | $22.3_{\pm1.86}$ | $108.5_{\pm0.00}$ | $125.9_{\pm1.61}$ | $30.2_{\pm2.35}$ | $122.9_{\pm0.00}$ | $19.6_{\pm0.03}$ | $22.75_{\pm1.02}$ | $43.62_{\pm0.00}$ |
| H2 | Iteration | $25.3_{\pm1.68}$ | $12.5_{\pm1.57}$ | $49.6_{\pm2.56}$ | $75.9_{\pm3.69}$ | $15.7_{\pm6.73}$ | $65.0_{\pm5.36}$ | $12.6_{\pm1.96}$ | $11.9_{\pm2.95}$ | $2.0_{\pm0.00}$ |
| | Time/s | $130.39_{\pm1.95}$ | $42.25_{\pm0.97}$ | $195.81_{\pm1.35}$ | $220.76_{\pm3.21}$ | $59.65_{\pm4.07}$ | $230.56_{\pm0.67}$ | $39.05_{\pm0.98}$ | $42.65_{\pm0.93}$ | $54.69_{\pm0.15}$ |
| H3 | Iteration | $23.1_{\pm2.53}$ | $12.2_{\pm1.26}$ | $49.3_{\pm2.86}$ | $75.0_{\pm4.65}$ | $15.2_{\pm5.37}$ | $64.0_{\pm5.37}$ | $11.8_{\pm1.83}$ | $10.2_{\pm3.67}$ | $2_{\pm0.00}$ |
| | Time/s | $202.3_{\pm2.83}$ | $65.4_{\pm0.83}$ | $289.5_{\pm1.58}$ | $331.6_{\pm3.97}$ | $82.77_{\pm4.62}$ | $330.65_{\pm0.68}$ | $63.9_{\pm0.76}$ | $65.5_{\pm2.69}$ | $63.3_{\pm0.12}$ |

*Compared results.* Considering the purpose of the experiment on H3 for example, Figs. 2-5 illustrate the measurement configurations selection, calibration accuracy, training curves and position errors, respectively. Table III records the calibration accuracy of M1-9 on H1-3. Table VI lists the time costs of algorithms M1-M9 on H1-3. From the above results, the significant findings are summarized:

a) **APMC evidently outperforms its peers in industrial robot calibration accuracy.** Compared with M1-7, M8's calibration accuracy is the highest. As shown in Fig. 3 and Table III, the RMSE, Mean and MAX of M8 on H3 are 0.552, 0.452 and 0.891 respectively, which are about 6.28%, 10.67% and 14.08% lower than M3's 0.589, 0.506 and 1.037, 24.90%. From Table III, similar results are encountered in H1-2. Thus, the APMC algorithm evidently outperforms seven state-of-the-art algorithms (M1-7) in calibration accuracy.

b) **MCS can effectively diminish the impact of measurement noise for further enhancing calibration accuracy.** As depicted in Fig. 2(a), with the increase of measurement configurations on D'1-3, the observability indexes all first increase and then tends to be flat at about 100 samples. Hence, the number of measurement configurations on D'1-3 are all defined as 100. Furthermore, Fig. 2(b) illustrates the best MCS through the DE algorithm optimizing $O_6$ on D'1-3. Subsequently, we identify the kinematic parameters errors based on the results of MCS. Form Table III and Fig. 3, M9's RMSE, Mean and MAX on H3 are 0.530, 0.412 and 0.816, which are about 8.88%, 8.85% and 8.42% lower than M8's 0.552, 0.452 and 0.891. On H1-2, similar conclusions are also encountered. Hence, MCS can effectively enhance calibration accuracy.

c) **AMPC algorithm converges fast.** As depicted in Fig. 4 and listed in Table VI, the convergence rate of M8 is fast. Compared with M1-7, it is converged involving the least iterations. The phenomenon's reason of it is that the AMPC algorithm can address the issue of the long tail convergence and thus enhance the converge rate.

d) **MCS-AMPC achieves high computational efficiency.** For instance, as illustrated in Table VI and Fig. 4(c), M9 consumes 63.2 seconds to converge on H3, which is 99.06% of 63.9 seconds by M7 and 96.64% of 65.5 seconds by M8. However, as depicted in Table VI, different outcomes are encountered on H1-2, which is attributed to the fact that the MCS of 3 planes can be computed in parallel. More significantly, after MCS, the number of computational samples for each plane is decreased, which can reduce the computational load and improve efficiency.

e) **The strategy of multi-planal constraints is more conducive to the kinematic parameters error identification.** As recorded in Table III, M9's RMSE, Mean and MAX on H3 (three planes) are about 8.55%, 8.85% and 7.90% lower than that on H1 (one plane), and about 4.91%, 3.06% and 1.92% lower than that on H2 (two planes). On M1-8, similar outcomes are also achieved. Hence, the results solidly support the performance of the strategy of multi-planal constraints.

f) **MCS-AMPC algorithm evidently narrows the robot's position errors.** The identification results are listed in Table V. Furthermore, we select 30 samples from 100 obtained by MCS to compare the performance of M1-9 on H3, as described in Fig. 5(a)-(c). The results show that after calibration, the position errors are evidently decreased and that M9 is the minimum among all algorithms. As listed in Table VI, on H1-2, similar conclusions are also encountered. Hence, M9, i.e., the proposed MCS-AMPC with the highest calibration accuracy is verified to be practically feasible without overfitting risk.

TABLE V. D-H PARAMETERS AFTER M9 CALIBRATION.

| Joint $i$ | $\alpha_i/°$ | $a_i/mm$ | $d_i/mm$ | $\theta_i/°$ |
|---|---|---|---|---|
| 1 | -90.1833 | 249.2389 | 653.8279 | 0.6879 |
| 2 | -0.5866 | 897.8359 | -0.4901 | -90.9871 |
| 3 | 90.7071 | -205.1962 | -0.1805 | 180.0269 |
| 4 | -90.0511 | 0.2569 | 1032.0231 | 0.3768 |
| 5 | 90.1526 | -0.1388 | 0.5856 | 90.2798 |
| 6 | 0.1601 | -0.5536 | 199.9666 | 0.9611 |

## V. CONCLUSIONS

Aiming at accurately calibrating the industrial robot, this paper proposes a novel MCS-AMPC algorithm. The MCS is employed to suppress measurement noise, and then the AMPC is adopted to identify the kinematic parameter errors of the industrial robot. The extensive empirical researches solidly support the high calibration performance of the MCS-AMPC. Hence, we develop a highly feasible calibration method for the industrial applications of robots. Presently, the following issues remain open:

a) GPU or other acceleration schemes is worthily invested in enhancing the compatibility of MCS.
b) To reduce the dynamic errors is also desired.